%% file: main.tex
\documentclass[10pt,twocolumn,letterpaper]{article}

\usepackage{iccv}
\usepackage{times}
\usepackage{epsfig}
\usepackage{graphicx}
\usepackage{amsmath}
\usepackage{amssymb}

\usepackage[pagebackref=true,breaklinks=true,letterpaper=true,colorlinks,bookmarks=false]{hyperref}

\newcommand{\mysubsubsection}[1]{\noindent\textbf{#1}}

\iccvfinalcopy 

\def\iccvPaperID{1463} 

\ificcvfinal\fi
\begin{document}

\title{Learning Temporal Action Proposals With Fewer Labels}
\author{Jingwei Ji, Kaidi Cao, Juan Carlos Niebles\\
Stanford University\\
{\tt\small \{jingweij, kaidicao, jniebles\}@cs.stanford.edu}
}

\maketitle

\begin{abstract}
  Temporal action proposals are a common module in action detection pipelines today. Most current methods for training action proposal modules rely on fully supervised approaches that require large amounts of annotated temporal action intervals in long video sequences. The large cost and effort in annotation that this entails motivate us to study the problem of training proposal modules with less supervision. In this work, we propose a semi-supervised learning algorithm specifically designed for training temporal action proposal networks. When only a small number of labels are available, our semi-supervised method generates significantly better proposals than the fully-supervised counterpart and other strong semi-supervised baselines. We validate our method on two challenging action detection video datasets, ActivityNet v1.3 and THUMOS14. We show that our semi-supervised approach consistently matches or outperforms the fully supervised state-of-the-art approaches.
\end{abstract}

\input{1-Introduction.tex}
\input{2-RelatedWork.tex}
\input{3-Model.tex}
\input{4-Experiments.tex}
\input{5-Conclusion.tex}

\mysubsubsection{Acknowledgments.} This work has been supported by Panasonic and JD.com American Technologies Corporation (``JD'') under the SAIL-JD AI Research Initiative. This article solely reflects the opinions and conclusions of its authors and not Panasonic, JD or any entity associated with Panasonic or JD.com.

{\small
\bibliographystyle{ieee_fullname}
\bibliography{egbib}
}

\end{document}

%% file: 1-Introduction.tex
\section{Introduction}
With millions of cameras in the world, a tremendous amount of videos are generated and transmitted every day. A very important subject in these videos is humans performing activities and actions. This has motivated the computer vision community to study algorithms for understanding actions from video collections. An important task for action understanding is action detection, or temporal action localization, where the goal is to temporally localize all actions of interest within long video sequences. A common approach to tackle this problem is to first generate \emph{temporal action proposals} to localize temporal intervals of interest, which are then fed into a classifier to obtain the corresponding action labels. In this paper, we focus on the temporal action proposal module.

To achieve high prediction accuracy, most of the existing state-of-the-art algorithms for temporal action proposals use supervised deep learning approaches \cite{sst, escorcia2016daps, gao2018ctap, bsn}. Such approaches require large amount of \textit{labeled} videos. Different from labeling in other vision tasks like image recognition, labeling temporal boundaries of actions in untrimmed videos is much more time-consuming. On the other hand are unsupervised learning approaches \cite{soomro2017unsupervised} where no label is needed for training. Although they are free from the burden of labeling, the overall performance in many tasks is usually inevitably poor than that of supervised approaches.

\input{Fig-Pull.tex}

Semi-supervised learning is a well fit solution when large amount of data is available but only a small portion is labeled. Different from unsupervised learning, semi-supervised learning still leverages labeled data as strong supervision for high prediction accuracy. Compared to supervised learning, semi-supervised learning is less likely to overfit on the small labeled dataset because it can make use of the unlabeled data. Semi-supervised learning has been effective in image classification \cite{tempensem, vat, ladder, meanteacher}, but has never been explored to assist generating temporal action proposals. In our problem setup (see Figure \ref{fig:teaser}), we assume that during training only a part of the videos come with temporal boundary labels of actions for supervised learning. In the meanwhile, other videos with no labels or annotations are available to be leveraged by the training process. By extending the knowledge extracted from the labeled set to the unlabeled set, we can obtain a more robust model due to the regularization role that the unlabeled data can play.

One core philosophy behind semi-supervised learning methods is to train the model with smooth and consistent classification boundaries that are robust to stochastic perturbation. To find a smooth manifold of data, Tarvainen \etal. \cite{meanteacher} proposes Mean Teacher which averages the ``student'' models at different training iterations into a ``teacher'' model. We embrace this architecture into our model design. To improve the robustness of the model, it is critical to introduce random perturbations on the input to the student model. In particular for the task of temporal action proposals in videos, the perturbations should be designed to benefit sequence learning. However, the prior work has not proposed appropriate perturbations for sequence data such as videos.

We propose two types of sequential perturbations: Time Warping and Time Masking. Time Warping is a resampling layer which distorts video sequences along the temporal dimension, providing perturbations for time-sensitive tasks like temporal action proposals. Time Masking randomly masks some frames of the input videos. During training, the masked student models only see parts of the videos while they are encouraged to predict the same boundaries as the unobstructed teacher model predicts. These sequential perturbations allow our optimized model to be more robust and generalize better to unseen data.

Our main contributions are as follows:
\textbf{(1)} To the best of our knowledge, we are the first to incorporate semi-supervised learning in temporal action proposals to achieve label efficiency.
\textbf{(2)} We have designed two essential types of sequential perturbations for this semi-supervised framework and validated them against strong semi-supervised baselines in key experiments of temporal action proposals.

%% file: Fig-Pull.tex
\begin{figure}[t]
\centering
\includegraphics[width=\linewidth]{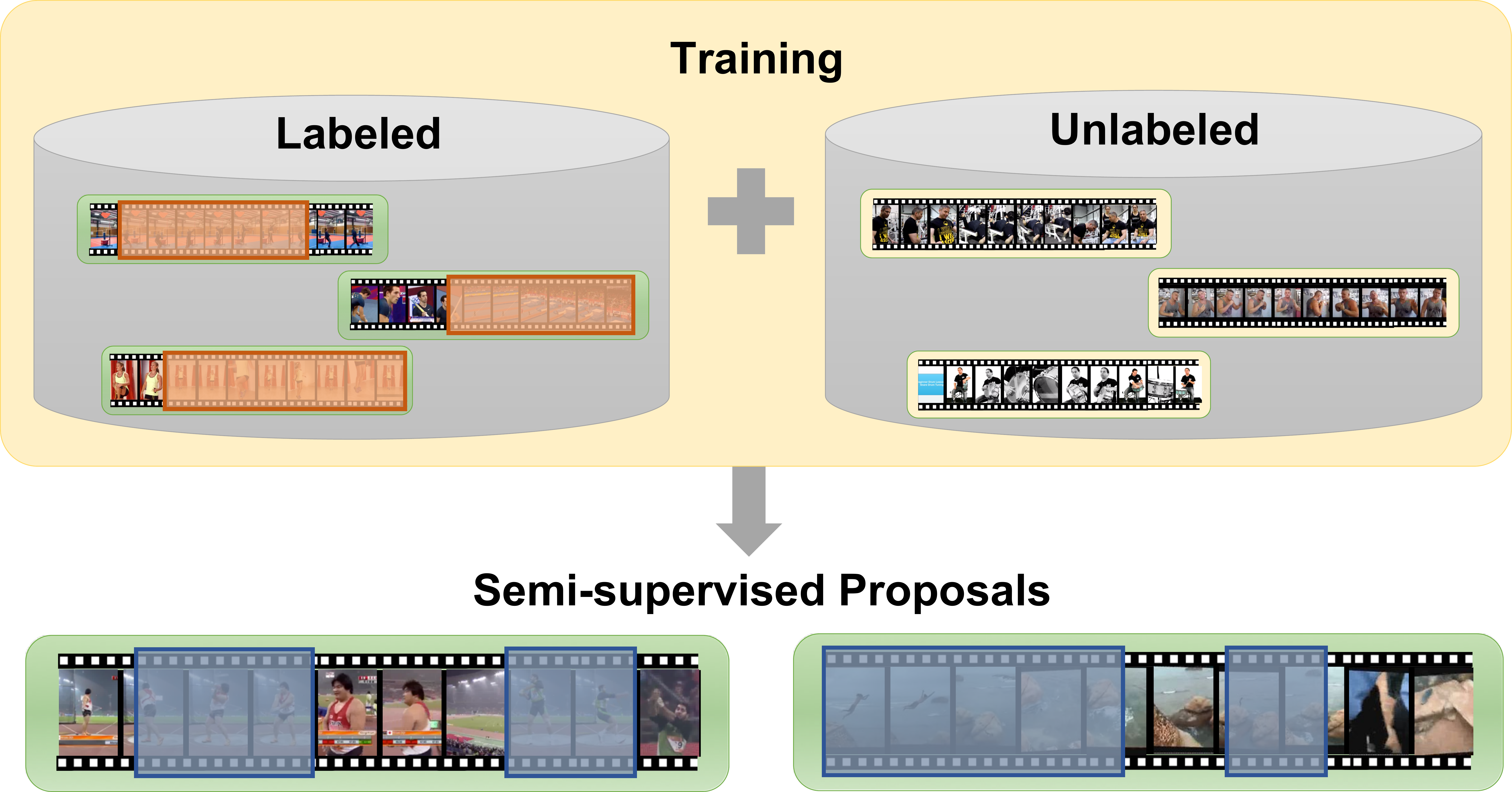}
\caption{With only a part of training videos labeled with ground truth proposals, our semi-supervised framework can generate temporal action proposals with better quality than the state-of-the-art fully-supervised approaches.}
\vspace{-1em}
\label{fig:teaser}
\end{figure}

%% file: 2-RelatedWork.tex
\section{Related Work}

\input{Fig-Arch.tex}
\noindent\textbf{Temporal Action Detection and Proposals.}
Given a long, untrimmed video, temporal action detection aims to localize each action instance with its start and end times as well as its action class \cite{caba2016fast, ganghua_tal, escorcia2016daps, gao2017turn, lin2017single, singh2016untrimmed, yuan2017temporal}.

Traditionally, many approaches address the problem by exhaustively applying an action classifier in a sliding windows fashion \cite{duchenne2009automatic, karaman2014fast, ni2016progressively, oneata2014lear, wang2014action, yuan2016temporal}. These methods are typically inefficient in terms of computation cost, since they need to cover temporal windows of different lengths at each location throughout the whole untrimmed video. 

Inspired by recent success in proposal-plus-classification approaches of image object detection, another group of two-stage methods first propose action-agnostic temporal segment in video, then classify the action of the trimmed clips. Buch \etal \cite{sst} propose a network that performs single-stream temporal action proposal generation, avoiding computation cost brought by sliding window. Shou \etal \cite{shou2016temporal} use 3D ConvNets to generate temporal proposals. There are also end-to-end frameworks that enable joint optimization of proposal generation and action classification. Buch \etal \cite{buch2017end} introduce semantics constraints for curriculum training in end-to-end temporal action localization. Chao \etal \cite{rethinking} adopt Faster R-CNN \cite{ren2015faster} for action localization task.

The proposals generated in the above methods are often dependent on pre-defined anchors, lacking flexibility and preciseness of temporal bounds. Instead, Zhao \etal \cite{zhao2017temporal} simplify the proposal generation problem into classifying the actionness of every short video snippet, post-processed by a watershed algorithm. Gao \etal \cite{gao2018ctap} and the Boundary Sensitive Network (BSN) \cite{bsn} further infer whether a video snippet is the start or end of an action to obtain more precise boundaries, in which the BSN has become the state-of-the-art on the temporal action proposal task on ActivityNet Challenge \cite{anet}. 

Previous research is dedicated to develop better action proposal models trained with labeled videos. In parallel, we explore how to utilize unlabeled videos to further improve proposal and detection performance. In this work, we focus on evaluating our semi-supervised framework with the BSN due to its superior performance, though our framework's flexibility allows it to be combined with other temporal action proposal architectures as well.

\noindent\textbf{Semi-supervised Deep Learning.}
Semi-supervised learning has a rich history that spans decades \cite{chapelle2009semi, zhu2003semi}. Instead of a comprehensive review, our focus is limited to semi-supervised deep learning. A common approach is to train a neural network by jointly optimizing a supervised classification loss on labeled data and an additional unsupervised loss on both labeled and unlabeled data \cite{tempensem, vat, ladder, meanteacher}. Consistency regularization has been widely used for the unsupervised loss, which encourages the model to generate consistent outputs when the raw inputs or intermediate feature maps are perturbed.

Here we summarize some examples of semi-supervised deep learning using consistency regularization. Ladder Networks \cite{ladder} incorporate a reconstruction branch as the unsupervised task; they enforce consistency losses between encoded and decoded activation maps at each training step. $\Pi$-Model \cite{tempensem} simplifies Ladder Networks and only imposes consistency loss between outputs with different perturbations on data. Next, Temporal Ensembling \cite{tempensem} applies a consistency loss to model outputs and a more stable target: the exponential moving average of model outputs at each epoch. Instead of averaging outputs, the more powerful Mean Teacher \cite{meanteacher} averages the weights of models at each training step (a.k.a. ``student" models) into a separate ``teacher" model, whose outputs serve as the target in the consistency loss. Orthogonal to the above approaches, Virtual Adversarial Training (VAT) \cite{vat} proposes using virtual adversarial noise instead of random noise as the data perturbation. In our work, we also impose consistency regularization between outputs of student and teacher models, and propose Time Warping and Time Masking as the data perturbations specifically for video data.

Semi-supervised learning has been applied to sequence learning as well. Dai \etal \cite{semisequence} propose a sequence autoencoder for text classification. Pr\'emont-Schwarz \etal \cite{recurladder} combine Ladder Networks with recurrent neural networks and evaluate their model on image classification on the Occluded Moving MNIST dataset. Clark \etal \cite{cvt} propose cross-view training for multiple language tasks. Miyato \etal \cite{vat_text} apply VAT \cite{vat} on text classification. Although not designed for video analysis, some of the above approaches \cite{cvt, recurladder} also embrace the idea of masking either on patches in images or words in sentences, and they inspire our Time Masking. 

There is also work on \textit{weakly supervised learning} for temporal action detection \cite{bojanowski2014weakly, chang2019d3tw, huang2016connectionist, shou2018autoloc}, which differs from our semi-supervised setting. In the weakly supervised temporal action detection, part of the training data are fully labeled with the temporal boundaries and action classes while the rest of data are annotated with ``weak" labels, either video-level classes or order lists of actions in the video. Instead, we do not assume availability of any kind of labels for the unlabeled videos used in our semi-supervised training, which entails a harder but more label-efficient task.

%% file: Fig-Arch.tex
\begin{figure*}[t]
\centering
\includegraphics[width=0.9\linewidth]{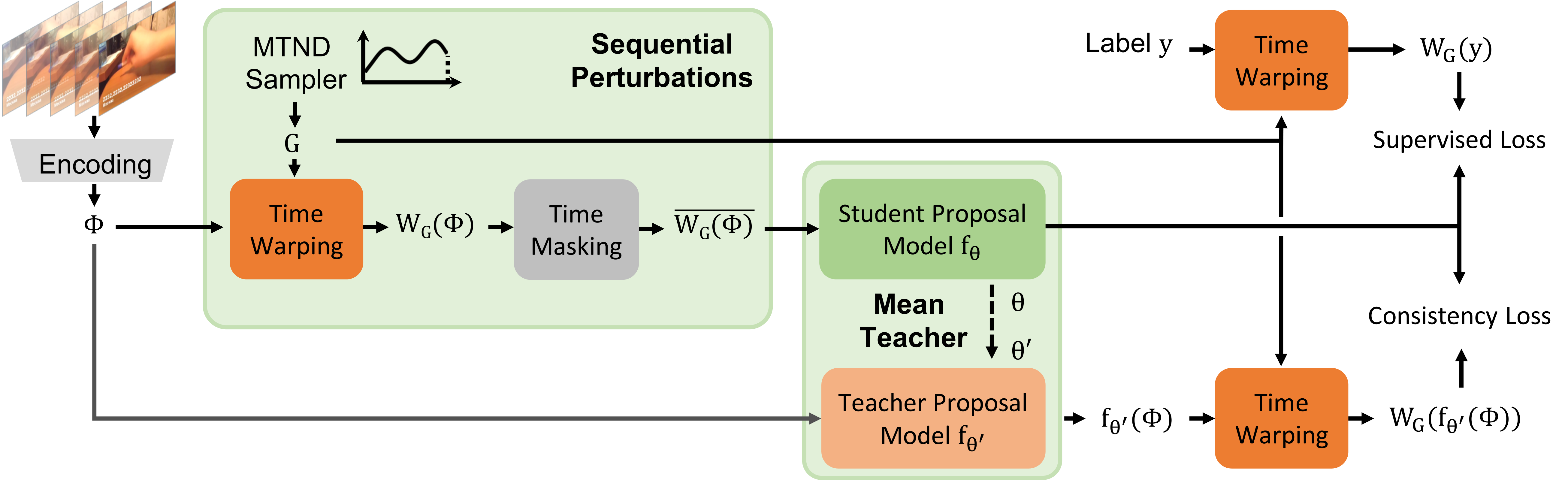}
\caption{Overview of our method. Given an untrimmed video as input, we first encode it into a feature sequence $\Phi$. Next, sequential perturbations including Time Warping and Time Masking are applied to $\Phi$ and the student proposal model takes this perturbed sequence as the input. Instead, the teacher model predicts directly on the unobstructed $\Phi$. In the end, the student model is jointly optimized with a supervised loss applied to labeled videos and a consistency loss to all videos.}
\label{fig:model}
\end{figure*}

%% file: 3-Model.tex
\section{Technical Approach}
Our main goal is to generate high-quality temporal action proposals with a relatively small amount of labels. This requires us to best utilize the labeled data with a powerful supervised proposal model while, at the same time, leveraging unlabeled data with an unsupervised auxiliary task designed for video understanding. Although our approach is agnostic to specific proposal methods, to validate the semi-supervised framework, we build our model on top of a state-of-the-art fully-supervised proposal generation network, the Boundary Sensitive Network \cite{bsn}. We extend the Mean Teacher framework \cite{meanteacher} with two types of sequential perturbations for training the proposal model: Time Warping and Time Masking. See Figure \ref{fig:model} as an overview of our method.

\input{Fig-TimeWarp.tex}

\subsection{Video Encoding}
The purpose of video encoding is to obtain a condensed video representation, which captures the appearance and motion patterns of a video. Given an untrimmed video with $N$ frames as the input, we first divide it into non-overlapping short snippets which contain $\delta$ frames each, forming a sequence of snippets $S=\{X_1, X_2, ..., X_T\}$, where $T=N/\delta$. As illustrated in prior work \cite{i3d,tsn}, both appearance and motion features contribute to action understanding, so we encode both the RGB frames and the optical flows of each video, then concatenate the encoded vectors. In particular, we use \cite{tsn} as the video encoder $\phi$ as in the fully-supervised baseline \cite{bsn}. The encoder generates a sequence of feature vectors $\Phi = \{\phi(X_1), \phi(X_2), ..., \phi(X_T)\} \in \mathbb{R}^{T \times D}$. Then we feed sequences of feature vectors into the following modules in mini-batches. Labeled and unlabeled videos share the same video encoder $\phi$ and they co-exist in the same mini-batch.

\subsection{Temporal Action Proposal Model}
Our semi-supervised model is sufficiently flexible that it can be built upon various fully-supervised temporal action proposal networks as long as they take sequential data as input. Specifically, we choose Boundary Sensitive Network (BSN) \cite{bsn}, a top performer in the temporal action proposal task in ActivityNet challenge 2018.

The same video encoding as in \cite{bsn} is performed as the first step, then $\Phi$ is directly fed into the BSN proposal model. The BSN is composed of a sequence of two trainable modules: a Temporal Evaluation Module (TEM) and a Proposal Evaluation Module (PEM). After the video encoding, TEM takes the snippet feature sequence $\Phi$ as the input. The sequence $\Phi$ is passed through temporal convolutional layers to generate three series of probability signals: actionness $p^a \in \mathbb{R}^{T}$, starting $p^s\in \mathbb{R}^{T}$ and ending $p^e \in \mathbb{R}^{T}$. Then proposals are generated according to these three signal sequences. Finally, PEM predicts a confidence score $p_{conf}$ for each proposal indicating how overlapped a proposal is with the closest ground truth interval, to decide if the proposal is accepted or rejected. Please refer to \cite{bsn} or our supplementary materials for more details of BSN.

\subsection{Mean Teacher Framework}

Now we introduce how we construct the semi-supervised learning framework for temporal action proposals. When only a small number of labeled training samples are available, deep models like BSN tend to over-fit and not able to extract enough knowledge from the training set to generalize to unseen videos. This can be mitigated by semi-supervised learning where unlabeled videos can also be used for training. Without ground truth labels, the supervised classification loss is undefined upon unlabeled videos. Instead, we need to introduce an unsupervised auxiliary task to leverage information from unlabeled videos.

As a baseline, we can directly adapt the Mean Teacher method on the temporal action proposal model to form the semi-supervised learning framework. In this framework, there are two models: a \textit{student} proposal model $f_{\theta}$ and a \textit{teacher} proposal model $f_{\theta '}$. The student learns as in fully-supervised learning, with its weights $\theta$ optimized by the supervised classification losses applied on labeled videos. The teacher proposal model has the identical neural network architecture as the student, while its weights $\theta'$ are generated by averaging $\theta$ from different iterations of training: 
{\small \begin{equation}
    \theta'_{i} = \alpha \theta'_{i-1} + (1 - \alpha) \theta_{i}
\end{equation}
}where $\alpha$ is a smoothing coefficient parameter and $i$ denotes the training iteration. As an ensembled model, the teacher embeds input snippet features into a smooth manifold and outputs more consistent predictions than students. Then the unsupervised task is to impose consistency regularization between the outputs from the student and the teacher model, with both labeled and unlabeled videos as input.

\subsection{Sequential Perturbations}
Beyond the Mean Teacher framework, stochastic perturbations have been found crucial for learning robust models by many semi-supervised learning works \cite{tempensem, vat, ladder, meanteacher}. A typical way of perturbation is adding noise to feature maps. Mean Teacher \cite{meanteacher} adds Gaussian noise to intermediate feature maps of both student and teacher models, whereas VAT \cite{vat} adds adversarial noise to the input. In video analysis, we further explore what other specific perturbations are necessary for sequential learning. We propose two sequential perturbations: Time Warping and Time Masking.

\mysubsubsection{Time Warping.}
Time Warping is essentially a resampling layer, which resamples a sequence of feature vectors $\Phi \in \mathbb{R}^{T \times D}$ along the time dimension guided by a randomly generated 1-D flow-field grid. Time Warping is vital for semi-supervised temporal action proposals: First, by propagating labels to unlabeled locations in the feature space, resampling leads to smoother predictions (Figure \ref{fig:timewarp} (a)); second, Time Warping serves as a way of data augmentation, providing more labeled data for training, which is especially helpful in the case when we have few labels; third, stretching and compressing input signals can generate more variants to learn in certain tasks, like temporal action proposals, which require accurate starting/ending location prediction.

To perform warping on the input feature sequence $\Phi$, each output feature vector is computed by applying linear sampling on $\Phi$ according to a dense 1-D grid $G = \{g_t\}$, where $g_t$ is the temporal location to sample an output feature vector. Critical in performing Time Warping, the grid should include long-term distortion which slows down some parts of the video while speeds up the other parts; it should also contain short-term stochastic noise. With these considerations, we propose a Mixed Truncated Normal Distribution (MTND) sampler (Figure \ref{fig:timewarp} (b)) to generate grids.

A MTND is formed by mixing $n$ truncated normal distributions $\mathcal{N}_{0}^{T}(\mu_i, \sigma_i), i \in \{1, 2, ..., n\}$ by different weights. Since we only want to interpolate the input sequence, the distribution is truncated at the starting (0) and ending ($T$) locations. The means $\mu_i$'s are sampled from a uniform distribution and the standard deviations $\sigma_i$'s are sampled from a log-uniform distribution. Given a MTND, we sample $T$ locations from it as the grid $G$, then we perform the warping and obtain $\mathcal{W}_G(\Phi) \in \mathbb{R}^{T \times D}$.

\mysubsubsection{Time Masking.}
Besides Time Warping, we propose a Time Masking operation as another source of sequential perturbations during training. In our pipeline, Time Masking follows Time Warping and takes $\mathcal{W}_G(\Phi)$ as input. The idea of Time Masking is simple: some snippets in the input sequence are masked out from the student model, while the teacher model can see the whole unobstructed video sequence. We denote the output of the Time Masking as $\overline{\mathcal{W}_G(\Phi)}$ . During the training, the masked student models at each iteration are encouraged to generate the same outputs as the teacher does, even though they could not access the entire information of input videos.

\input{Fig-TimeMask.tex}

Time Masking can be viewed as a special Dropout layer (Figure \ref{fig:timemask}). In the regular Dropout layer, the neurons in one snippet are not likely to be entirely dropped, which gives the model a chance to peek some information from every snippet in the receptive field. Instead, in Time Masking, no information of the dropped snippet will be passed to the next layers. The student model will be forced to aggregate information from temporal context to make prediction on dropped snippets. Such capability of temporal context aggregation will be learned both from supervised losses on the labeled videos and the consistency with the teacher model on all training data.

\subsection{Training}
\label{subsec:train}
Training our semi-supervised framework includes two parts: minimizing the supervised losses on labeled data and the consistency loss on all training data. Although we have student and teacher models, only weights in student models are optimized via back-propagation, and weights in the teacher model are the averaged weights of students.

\noindent\textbf{Supervised Losses.}
Aligned with the fully-supervised proposal model, our semi-supervised framework uses the same supervised losses for training as in BSN. Please refer to \cite{bsn} or our supplementary materials for details of the losses. In our semi-supervised framework, the output of the student proposal model corresponds to the sequential input distorted by Time Warping. Thus the labels $y$ also need to be resampled according to the same grid generated by the MTND sampler. With the warped labels $\mathcal{W}_G(y)$, we enforce the supervised losses on the student output $f_{\theta}(\overline{\mathcal{W}_G(\Phi)})$. Note that the supervised losses can only be applied on labeled videos in the training set.

\noindent\textbf{Consistency Regularization.}
The consistency loss treats the outputs of the teacher model as labels and encourages the student to learn a smooth manifold like the teacher's. Unlike the supervised losses, the consistency loss can be applied to both labeled and unlabeled videos in the training set. Similar to how we handle the labels in supervised losses, we also warp the outputs of the teacher to be $\mathcal{W}_G(f_{\theta'}(\Phi))$. The consistency loss then measures the distance between the student outputs and the warped teacher outputs:
{\small \begin{equation}
    L_{cons} = D(f_{\theta}(\overline{\mathcal{W}_G(\Phi)}), \mathcal{W}_G(f_{\theta'}(\Phi)))
\end{equation}
}For the distance function $D$, we use Mean Squared Error in all experiments. Same as the supervised optimization, only weights in the student model are trained. The consistency loss and the supervised losses are summed as the total loss.

%% file: Fig-TimeWarp.tex
\begin{figure*}[t]
\centering
\includegraphics[width=0.8\linewidth]{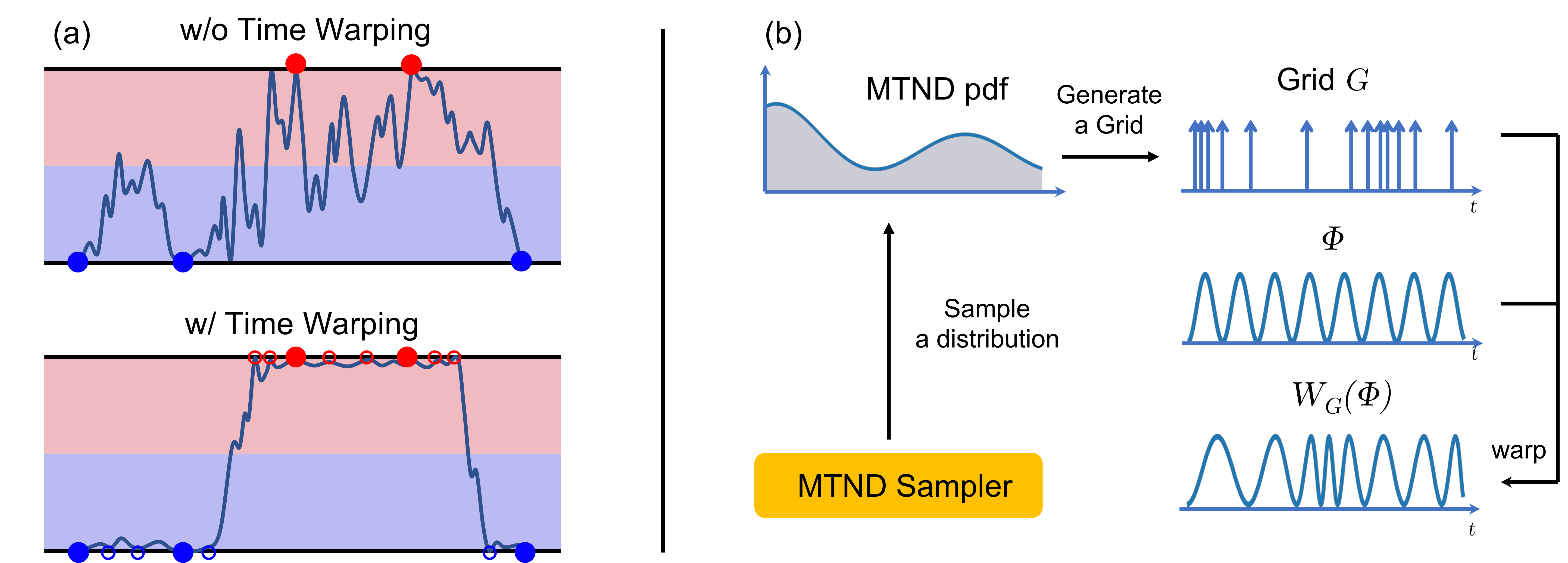}
\caption{Time Warping. (a) With Time Warping, we can sample more snippet features in the encoded space. Here we show a simple example of binary classification for each snippet feature (dimension reduced to 1). Resampling new feature points (the empty circles) among labeled snippet features (the filled circles) encourages the student model to generate a smoother manifold for prediction.(b) To perform Time Warping, we first sample a mixed truncated normal distribution for generating the 1-D grid $G$. Then we apply grid sampling on the feature sequence $\Phi$ to augment the data for training.}
\vspace{-1em}
\label{fig:timewarp}
\end{figure*}

%% file: Fig-TimeMask.tex
\begin{figure}[t]
\centering
\includegraphics[width=0.8\linewidth]{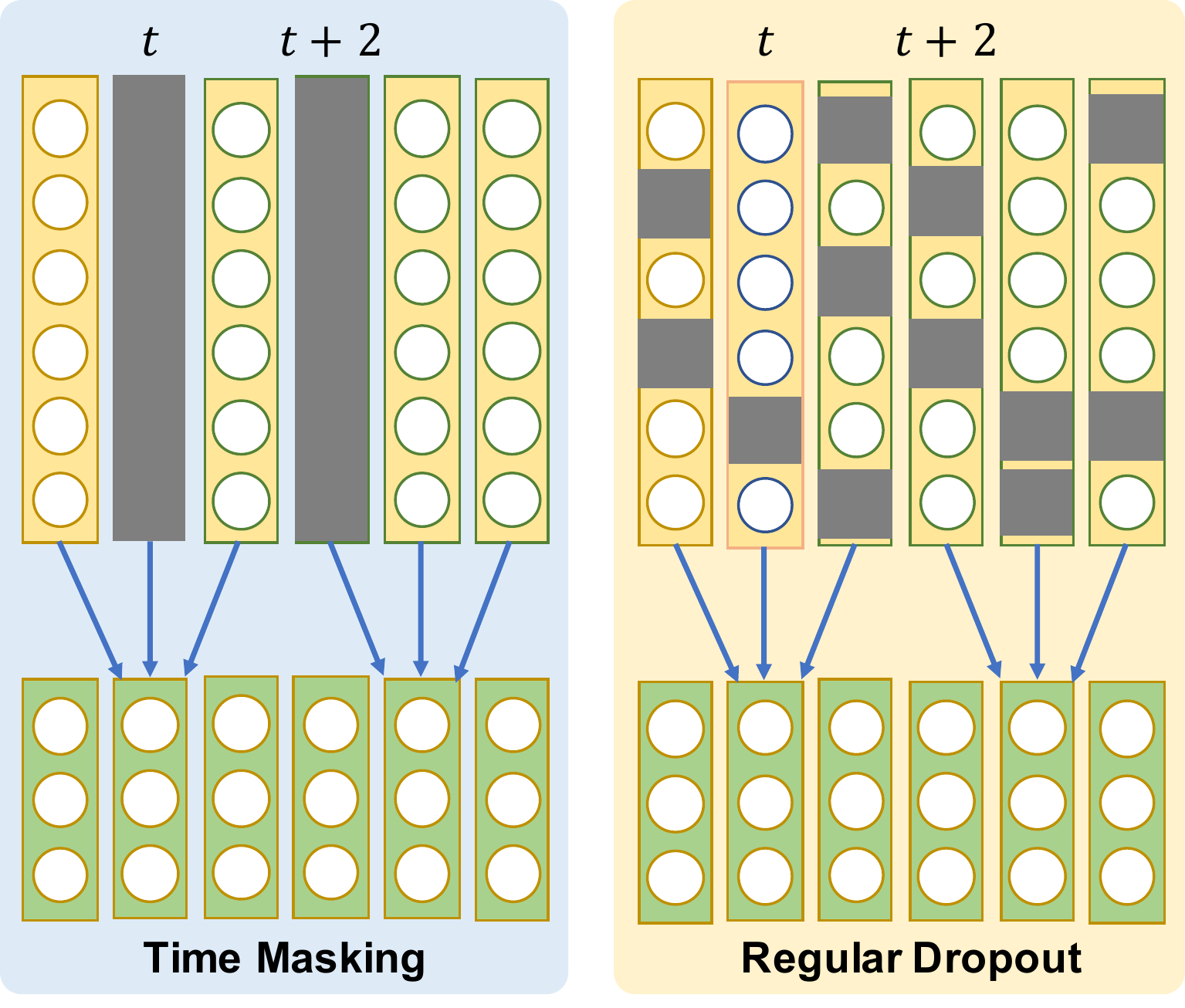}
\caption{Time Masking. Unlike dropout which randomly zero-outs some of the neurons in the input, Time Masking drops the entire feature vectors from the randomly selected time steps.}
\vspace{-1em}
\label{fig:timemask}
\end{figure}

%% file: 4-Experiments.tex
\section{Experiments}
\mysubsubsection{Datasets.}
We use ActivityNet v1.3 and THUMOS14 for all experiments. \textbf{ActivityNet v1.3} \cite{anet} is a large database for temporal action proposals and detection. It contains 19,994 videos of 200 activity classes and has been used in the ActivityNet Challenge 2016 to 2019. ActivityNet v1.3 is divided into training, validation and testing sets by a ratio of 2:1:1, and temporal boundaries of action instances are annotated in all videos. \textbf{THUMOS14} \cite{thumos} contains 200 and 213 temporal annotated untrimmed videos with 20 action classes in validation and testing sets, separately. The training set of THUMOS14 is the UCF-101 \cite{ucf101}, which contains trimmed videos for action classification task. Instead of training on these trimmed videos, we train our model on the untrimmed videos in validation set, and report performance on the test set.

\mysubsubsection{Evaluation Metrics.}
We evaluate our method on two tasks: temporal action proposals and temporal action localization. For proposals, we report Average Recall (AR) at various Average Number of Proposals per Video (AN). AR is defined as the average of all recall values with tIoU thresholds from 0.5 to 1 with a step size of 0.05. On ActivityNet v1.3, area under the AR vs. AN curve (AUC) is also used as a measurement, where AN varies from 0 to 100. For action localization, we calculate mean Average Precision (mAP) with different tIoU thresholds.

\mysubsubsection{Implementation Details.}
We follow the same pre- and post-processing as the BSN \cite{bsn}, including parameters used in Soft-NMS. For feature extraction on ActivityNet v1.3, we use the two-stream network \cite{tsn} pre-trained on Kinetics \cite{kay2017kinetics}. Different from the BSN's setting, our features are not pre-trained on ActivityNet classification task to avoid using extra labels which will contaminate the semi-supervised setup. We use the same video features as the BSN for all THUMOS14 experiments. For the semi-supervised training, we use EMA decay $\alpha=0.999$. Masking probability in Time Masking is fixed to 0.3.

\subsection{Temporal Action Proposals}

Taking a long, untrimmed video as input, our method aims to generate temporal boundaries determining the starting and ending time of each action instances. In this section, we compare the temporal action proposals generated by our model on ActivityNet v1.3 and THUMOS14 with fully-supervised BSN and other state-of-the-art methods to verify the effectiveness of our semi-supervised framework.

\mysubsubsection{Comparison to fully-supervised methods.} We first compare the action proposal results on ActivityNet-1.3 validation set under two training setups: (1) Our semi-supervised framework, where $x$ percent of training videos are labeled with temporal boundaries and $100-x$ percent of training videos are not; (2) State-of-the-art fully-supervised learning, where the same amount of labeled videos are used for training while no other data are used. With this comparison, we can see how our semi-supervised framework performs against the fully-supervised counterpart under different labeled/unlabeled ratio. 

To validate the label efficiency of our method, we vary the amount of labels for training, then measure the AUC and AR@100 of proposals generated by our method and the original BSN (Figure \ref{fig:anetlabel}). With only a part of the training set labeled, our method outperforms the fully-supervised baseline consistently under different ratio of labeled training videos. Notably, with only 60\% of the videos labeled, our semi-supervised model outperforms the state-of-the-art fully-supervised BSN trained with all labels in both metrics of AUC and AR@100 (Table \ref{tab:anet_prop}). Similarly, we examine the label efficiency on THUMOS14 (Figure \ref{fig:th14label}), and observe consistent superior performance as well.

\input{Fig-ANetLabelEffi.tex}

\input{Tab-1-ANetProp.tex}

\input{Fig-TH14LabelEffi.tex}

 We then compare the proposal generation on THUMOS14 with strong baseline models. Table \ref{tab:th14_prop} shows the comparison measured by average recall at various average number of proposals per video. Again, we outperform the BSN when trained with only 60\% of labels. Moreover, when 100\% of labels are available, our framework can further increase the average recalls.

\mysubsubsection{Comparison to semi-supervised baselines.}
Next, we investigate the performance of our framework against multiple semi-supervised baselines on THUMOS14 proposals with 60\% labels for training (Table \ref{tab:comparesemi}). We first implement and evaluate VAT \cite{vat} combined with BSN. The key idea of VAT is to improve model robustness to the approximately worst case perturbations instead of random ones. Similar to the VAT application to text classification \cite{vat_text}, we apply the adversarial noise to each video snippet embeddings, rather than directly to the raw input. VAT does not improve average recall by much, partly because that the worst case perturbations on video snippet embeddings are not significantly different to random noises.

\input{Tab-2-TH14Prop.tex}

\input{Tab-3-CompareSemi.tex}

We also investigate different variants of Mean Teacher \cite{meanteacher}. The vanilla Mean Teacher with only random noises and no dropout layers outperforms VAT. Also, adding VAT to Mean Teacher does not help much on better proposals. Mean Teacher with regular dropout further improves the quality of proposals, but not as powerful as our approach with Time Masking. With the same dropout/masking probability, although the regular dropout zeros the same amount of neurons as Time Masking per training step, it formulates an easier task for student models to learn since the student can rely on more snippets to do inference.

Finally, we examine the contributions of the two proposed sequential perturbations by removing them respectively. Both of them contribute to the proposals while Time Warping appears to play a major role.

\mysubsubsection{Qualitative Results.}
We visualize some temporal action proposals generated by our semi-supervised approach. Figure \ref{fig:qual} shows that our approach is able to generate more precise temporal boundaries than the fully-supervised baseline on THUMOS14 when both are trained with 60\% of labels.

\input{Fig-Ablation.tex}

\input{Tab-4-Det.tex}

\input{Fig-Qual.tex}

\subsection{Ablation Experiments}

To assess the functionality of the two proposed sequential perturbations, we run experiments on THUMOS14 with 60\% of the labels with different hyper-parameters used in Time Warping and Time Masking.

\mysubsubsection{Degree of distortion in Time Warping.}
The effect of Time Warping depends on the grid sampled from MTND sampler. Varying the number of truncated normal distributions and their scales, the MTND can go from a nearly uniform distribution to a very uneven one which will greatly distort the input sequence. We examine the impact of different degree of distortion in Time Warping on generated proposals. The degree of distortion is measured by the KL-divergence $D_{\text{KL}}(P \parallel Q)$ between the sampled MTND as $P$ and a uniform distribution as $Q$. Figure \ref{fig:ablation} (a) shows a sweet spot with $D_{\text{KL}}$ at an order of magnitude of $0.01$. When $D_{\text{KL}}$ approaches to 0, the effect of Time Warping diminishes; when the degree of distortion is too large, many parts of videos can hardly get sampled, equivalently decreasing the number of labels for training.

\mysubsubsection{Masking probability in Time Masking.}
We experiment with different probabilities of zeroing feature vectors in the sequence fed to Time Masking. As shown in Figure \ref{fig:ablation} (b), $p=0.3$ appears to be an optimal operating point, bringing appropriate difficulty to the students. Thus we fix this masking probability in all our experiments.

\subsection{Temporal Action Localization}

The end goal of generating temporal action proposal is temporal action localization, so we further evaluate our proposals for the localization task on THUMOS14. We follow the proposal-plus-classification two-stage approach as in \cite{sst, gao2017turn, bsn}. As the BSN does, we use the top-2 video-level classes predicted by UntrimmedNet \cite{singh2016untrimmed} on top of the proposals generated by different approaches. We report the mean Average Precision at different temporal IoU thresholds with 200 proposals per video on THUMOS14 (Table \ref{tab:det}). The direct comparison is with the fully-supervised BSN trained with all labels, where we achieve better performance on different temporal IoU thresholds from 0.3 to 0.7. When trained with all labels, our model further improves performance on action localization.

%% file: Fig-ANetLabelEffi.tex
\begin{figure}[t]
\centering
\includegraphics[width=\linewidth]{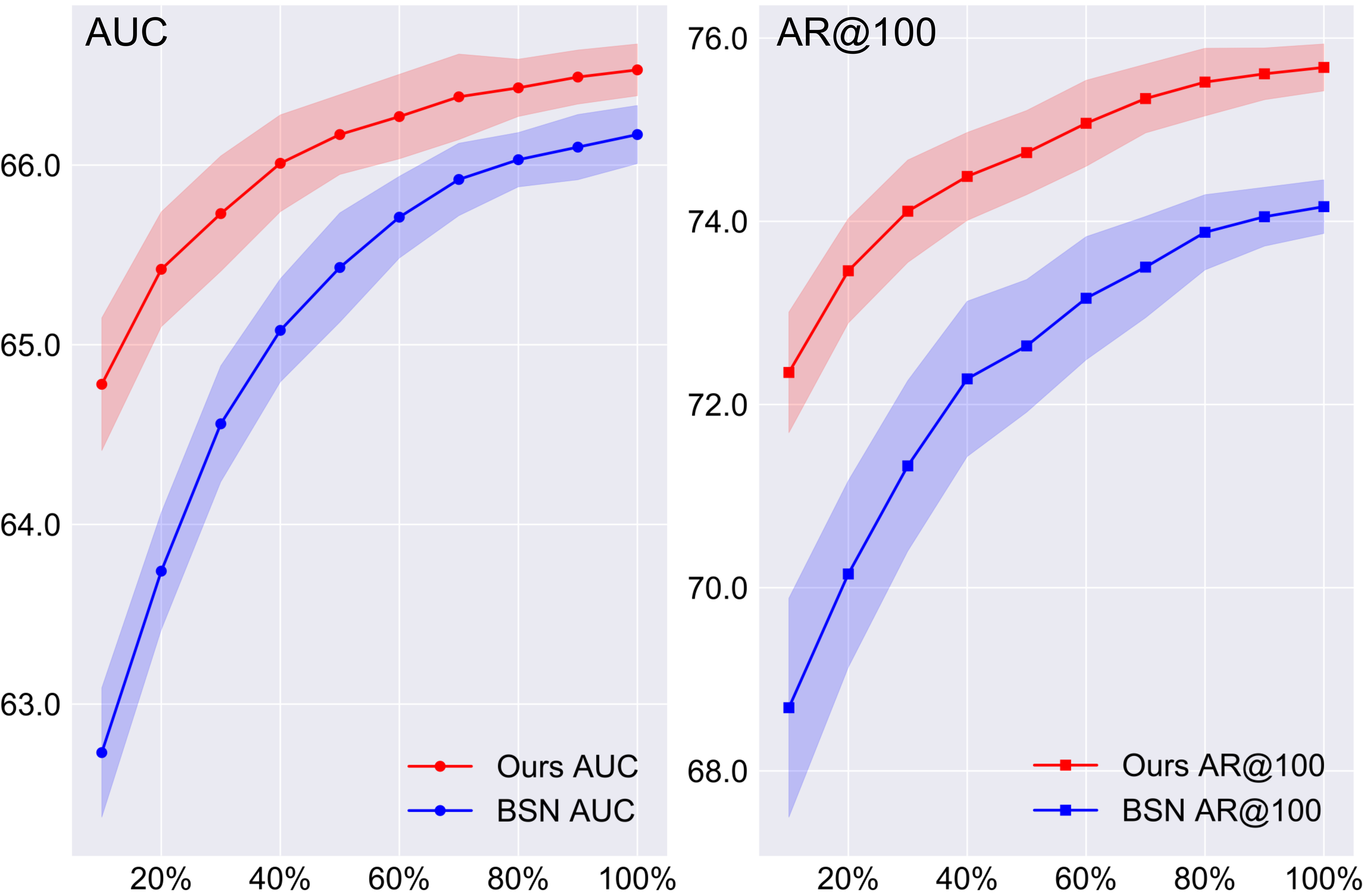}
\caption{Label efficiency experiments on ActivityNet v1.3. Varying the percentages of labels for training, we compare the AUC and AR@100 of the proposals generated by our semi-supervised method and the fully-supervised BSN counterpart.}
\label{fig:anetlabel}
\end{figure}

%% file: Tab-1-ANetProp.tex
\begin{table}[t]
\small
\begin{center}
\begin{tabular}{l|ccccc}
\hline
Method & SSN\cite{zhao2017temporal}   & CTAP\cite{gao2018ctap} & BSN\cite{bsn}   & Ours@60\%    \\ \hline
AR@100 & 63.52  & 73.17 & 74.16 & \textbf{75.07} \\ \hline
AUC    & 53.02  & 65.72 & 66.17 & \textbf{66.35} \\ \hline
\end{tabular}
\end{center}
\vspace{-1mm}
\caption{Comparison between our method and other state-of-the-art proposal generation methods on ActivityNet v1.3 in terms of AR@100 and AUC. We outperform all other methods while using only 60\% of the labels.}
\label{tab:anet_prop}
\end{table}

%% file: Fig-TH14LabelEffi.tex
\begin{figure}[t]
\centering
\includegraphics[width=\linewidth]{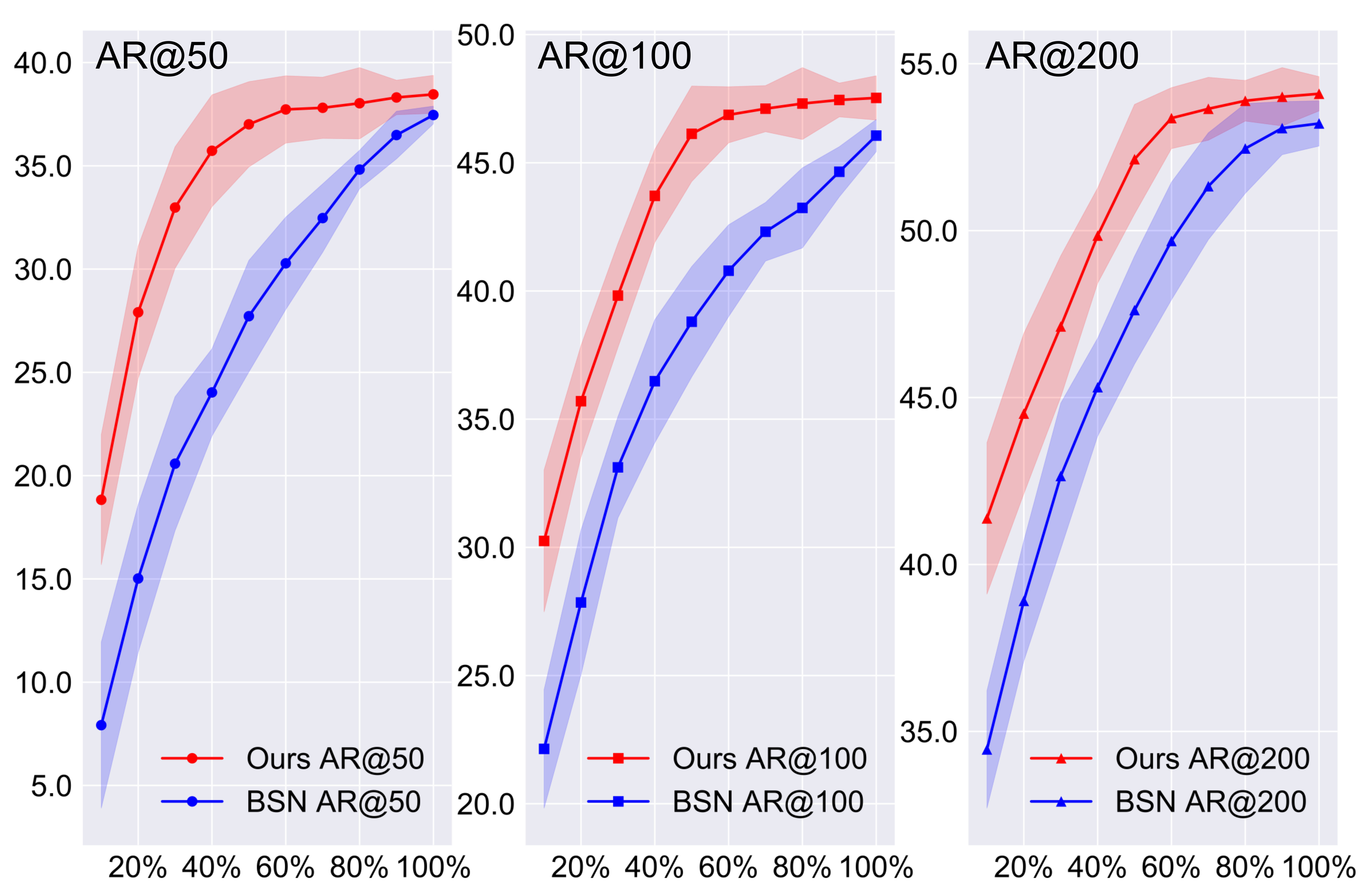}
\caption{Label efficiency experiments on THUMOS14. We report AR@50, @100, and @200 of the proposals generated by our method and the vanilla BSN when trained with different percentages of labels in the training set.}
\vspace{-1em}
\label{fig:th14label}
\end{figure}

%% file: Tab-2-TH14Prop.tex
\begin{table}[t]
\small
\begin{center}
\begin{tabular}{ll|lll}
\hline
Feature  & Method    & @50            & @100           & @200          \\ \hline
C3D      & DAPs \cite{escorcia2016daps}     & 13.56          & 23.83          & 33.96         \\
C3D      & SCNN-prop \cite{shou2016temporal} & 17.22          & 26.17          & 37.01         \\
C3D      & SST \cite{sst}      & 19.90           & 28.36          & 37.90          \\
C3D      & TURN  \cite{gao2017turn}     & 19.63          & 27.96          & 38.34         \\
C3D      & BSN \cite{bsn}      & 29.58          & 37.38          & 45.55         \\ \hline
2-Stream & TAG \cite{zhao2017temporal}     & 18.55          & 29.00             & 39.61         \\
Flow     & TURN \cite{gao2017turn}     & 21.86          & 31.89          & 43.02         \\
2-Stream & CTAP \cite{gao2018ctap}     & 31.03          & 40.23          & 50.13         \\
2-Stream & BSN@60\%  \cite{bsn}    & 30.28          & 40.79          & 49.03 \\
2-Stream & BSN@100\%  \cite{bsn}     & 37.46          & 46.06          & 53.21         \\ \hline
2-Stream & Ours@60\%     & \textbf{37.73} & \textbf{46.87} & \textbf{53.37} \\
2-Stream & Ours@100\%     & \textbf{38.46} & \textbf{47.53} & \textbf{54.10} \\ \hline
\end{tabular}
\end{center}
\caption{Comparison between our method and other state-of-the-art proposal generation methods on THUMOS14 in terms of AR@50, AR@100 and AR@200.}
\label{tab:th14_prop}
\end{table}

%% file: Tab-3-CompareSemi.tex
\begin{table}[t]
\small
\begin{center}
\begin{tabular}{l|lllll}
\hline
AN                     & @50            & @100           & @200          & @500           & \footnotesize @1000         \\ \hline
Vanilla BSN  & 30.28          & 40.79         & 49.03         & 57.58         & 62.35         \\ \hline
VAT \cite{vat}                    & 32.48          & 43.13          & 49.18         & 57.61          & 62.49         \\ \hline
MT \cite{meanteacher}          & 35.61          & 44.20          & 51.51         & 58.66          & 62.55         \\ 
MT + VAT & 35.63          & 44.21          & 51.49         & 58.64          & 62.56         \\ 
MT + Dropout & 35.73          & 44.25          & 51.56         & 58.67          & 62.58         \\ \hline
Ours -TW & 36.31          & 44.79          & 52.30         & 58.97          & 62.82         \\
Ours -TM & 37.24          & 45.37          & 52.65         & 59.74          & 63.10         \\
Ours                   & \textbf{37.73} & \textbf{46.87} & \textbf{53.37} & \textbf{60.81} & \textbf{64.59} \\ \hline
\end{tabular}
\end{center}
\caption{Comparison between fully-supervised and semi-supervised baselines trained with 60\% of the labels. We report AR at various AN on THUMOS14. Abbreviations: VAT for Virtual Adversarial Training, MT for Mean Teacher, TW for Time Warping, and TM for Time Masking. Our full model outperforms strong semi-supervised baselines.}
\vspace{-1em}
\label{tab:comparesemi}
\end{table}

%% file: Fig-Ablation.tex
\begin{figure}[t]
\centering
\includegraphics[width=\linewidth]{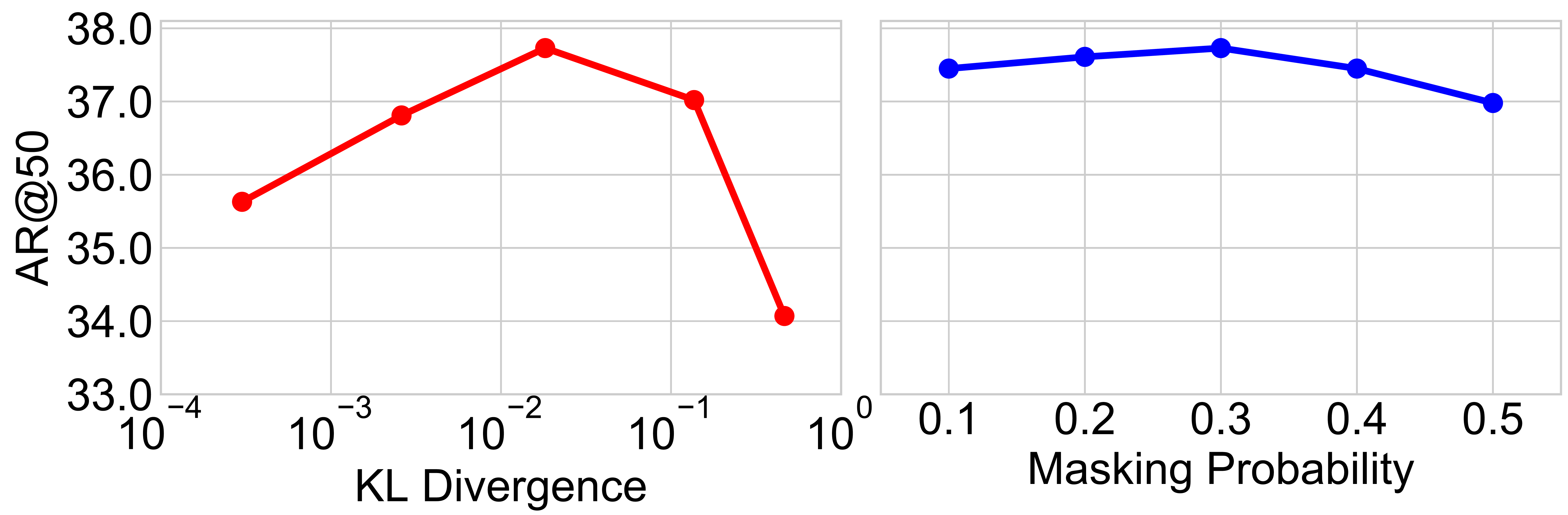}
\caption{Ablation experiments. We assess the effects of Time Warping and Time Masking under different hyper-parameter choices to find sweet points for better performance.}
\label{fig:ablation}
\end{figure}

%% file: Tab-4-Det.tex
\begin{table}[bt]
\small
\begin{center}
\begin{tabular}{l|rrrrr}
\hline
Method             & 0.7           & 0.6           & 0.5           & 0.4           & 0.3           \\ \hline
SST \cite{sst} + UNet                   & 4.7           & 10.9          & 20.0            & 31.5          & 41.2          \\
TURN \cite{gao2017turn} + UNet                  & 6.3           & 14.1          & 24.5          & 35.3          & 46.3          \\
BSN \cite{bsn} + UNet                   & 20.0            & 28.4          & 36.9          & 45.0            & 53.3          \\ \hline
Ours@60\% + UNet                  & \textbf{20.5} & \textbf{29.5} & \textbf{37.2} & \textbf{45.2} & \textbf{53.4} \\
Ours@100\% + UNet  & \textbf{20.7} & \textbf{29.9} & \textbf{37.9} & \textbf{46.3} & \textbf{55.1} \\ \hline
\end{tabular}
\end{center}
\caption{Action detection results on the testing set of THUMOS14 in terms of mAP@tIoU. We compare with proposal + classification methods, where classification results are generated by UntrimmedNet \cite{singh2016untrimmed}.}
\vspace{-1em}
\label{tab:det}
\end{table}

%% file: Fig-Qual.tex
\begin{figure*}[t]
\centering
\includegraphics[width=0.8\linewidth]{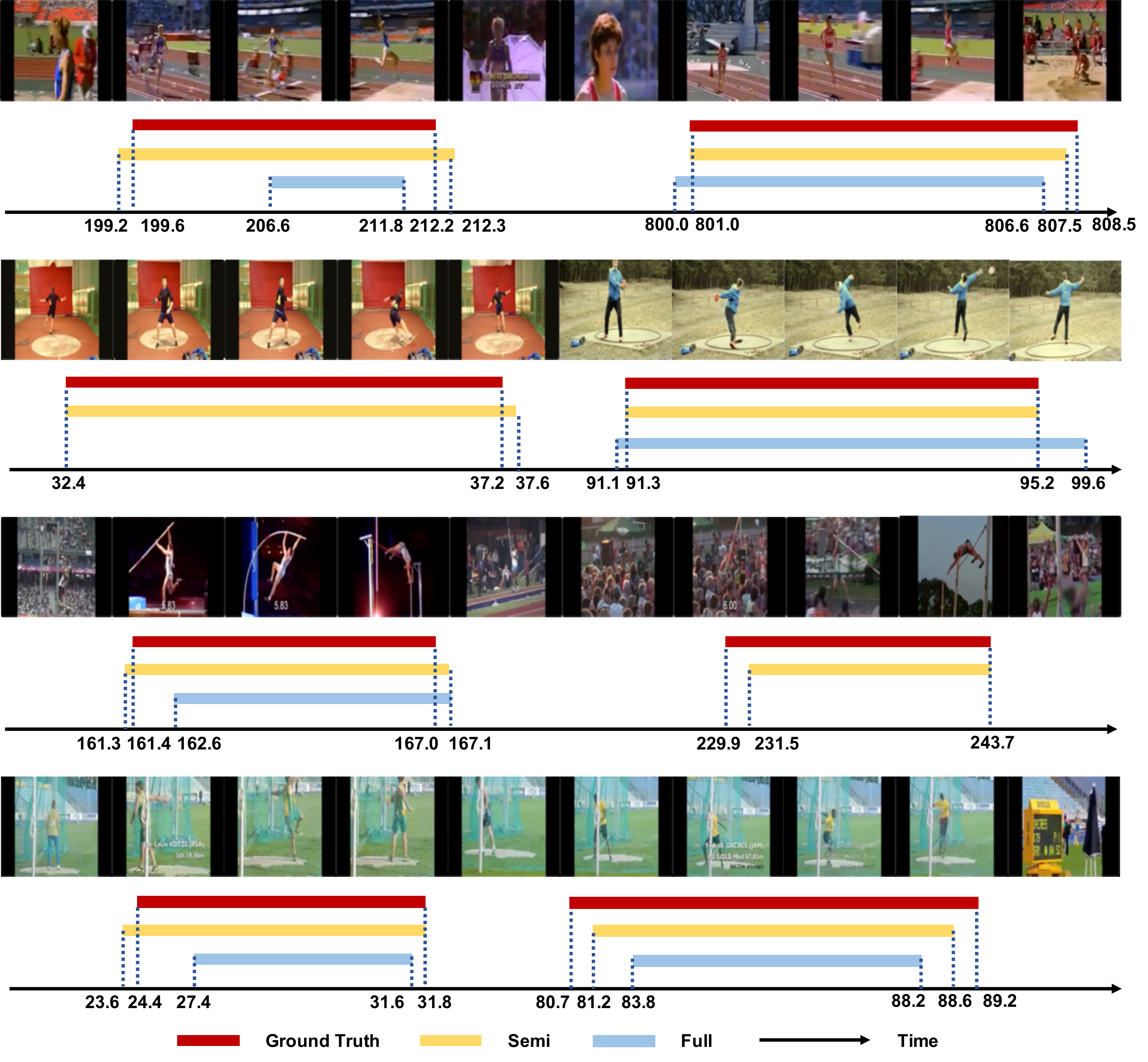}
\caption{
We compare THUMOS14 proposals generated by our semi-supervised method with the fully-supervised BSN trained using 60\% of the labels. We also show ground truth intervals for reference.}
\vspace{-1em}
\label{fig:qual}
\end{figure*}

%% file: 5-Conclusion.tex
\section{Conclusion}
We show that temporal proposal models can be trained with higher label efficiency by adopting our semi-supervised approach to learn their parameters. Our semi-supervised framework extends the Mean Teacher model with two proposed sequential perturbations for video understanding. We show empirically that our model achieves similar performance as the fully-supervised approach when trained with only 60\% of the labels, outperforming other semi-supervised baselines as well. Furthermore, we show that our semi-supervised proposals can be effectively applied to the problem of temporal action localization.